\documentclass[conference]{IEEEtran}

\ifCLASSINFOpdf

\fi

\usepackage{amsmath}
\usepackage{amssymb}
\usepackage{graphicx}
\usepackage{algorithm}
\usepackage[noend]{algpseudocode}
\usepackage{rotating,multirow}
\usepackage{amsmath,amssymb,mathtools,bm,etoolbox}
\usepackage{hyperref}
\usepackage{float}
\usepackage[caption=false]{subfig}
\usepackage{natbib}

\begin{document}

\title{Predicting the success of Gradient Descent for a particular Dataset-Architecture-Initialization (DAI) }

\author{\IEEEauthorblockN{\textbf{Umangi Jain}}
\IEEEauthorblockA{Department of Electrical Engineering\\
Indian Institute of Technology Madras\\
Chennai, 600036\\
ee16b124@smail.iitm.ac.in}
\and

\IEEEauthorblockN{\textbf{Harish G. Ramaswamy}}
\IEEEauthorblockA{Department of Computer Science\\
Indian Institute of Technology Madras\\
Chennai, 600036\\
hariguru@cse.iitm.ac.in}
}

\maketitle

\begin{abstract}
Despite their massive success, training successful deep neural networks still largely relies on experimentally choosing an architecture, hyper-parameters, initialization, and training mechanism. In this work, we focus on determining the success of standard gradient descent method for training deep neural networks on a specified dataset, architecture, and initialization (DAI) combination. Through extensive systematic experiments, we show that the evolution of singular values of the matrix obtained from the hidden layers of a DNN can aid in determining the success of gradient descent technique to train a DAI, even in the absence of validation labels in the supervised learning paradigm. This phenomenon can facilitate early give-up, stopping the training of neural networks which are predicted to not generalize well, early in the training process. Our experimentation across multiple datasets, architectures, and initializations reveals that the proposed scores can more accurately predict the success of a DAI than simply relying on the validation accuracy at earlier epochs to make a judgment. 
\end{abstract}

% \textit{Index terms: } 

\IEEEpeerreviewmaketitle

\section{Introduction}\label{s1}

% Motivation outline: empirical choice of init+arch+training scheme combined with ficklness ->  motivation to save compute time power/time -> early give-up -> scores -> bandit

Neural architectures have achieved state-of-the-art performance on various large-scale tasks in several domains spanning language, vision, speech, recommendations. However, modern deep learning algorithms consume enormous energy in terms of compute power [\cite{Li2016EvaluatingGPUs}, \cite{Strubell2019EnergyNLP}]. The abundance of data and improvements on the hardware side have made it possible to train larger models, increasing the energy consumption at a massive rate. These deep learning algorithms go through intense mathematical calculation during forward and backward pass for each piece of data to update the large weight matrices. Another crucial factor which significantly adds up to the compute power is the immense experimentation and tuning required to choose optimal initialization, architecture, and hyper-parameters. Several variants of a model are generated and trained for a particular task on a particular dataset to make an optimal choice.

While there is active research going on to make deep learning algorithms greener and economical like compressing the model [\cite{Kozlov2020NeuralInference}], training on a subset of the dataset [\cite{Frankle2018TheNetworks}],  designing hyper-efficient network [\cite{Tang2020SearchingConvolution}], there's still no standard mechanism to choose an initialization or architecture design. Though various theories have been proposed for choosing an initialization distribution, architecture design, training mechanism,  these nets are still most often manually designed through extensive experimentation.  
For instance, initialization schemes, such as Xavier [\cite{GlorotUnderstandingNetworks}], He [\cite{He2015DelvingClassification}], 
random orthogonal [\cite{Saxe2013ExactNetworks}] have been proposed in the past. Despite some theoretical backing, these commonly used initialization techniques are simple and heuristic [\cite{GoodfellowDeepLearning}]. The more recent neural architecture search (NAS) research has fueled efforts in automatically finding good architectures [\cite{Liu2018DARTS:Search}, \cite{Zoph2016NeuralLearning}]. It has been observed that these NAS algorithms are computationally very intensive and time-consuming as well. In addition, changing the dataset or tasks requires starting these algorithms anew. 

Furthermore, DNNs are characterized by variability in performance. The performance of the same architecture, training scheme, and initialization on a particular dataset can vary due to the stochastic nature of deep learning algorithm, system architecture, operating systems, and/or libraries.  Consequently, experimentation is required for choosing an optimal initialization weight matrix when sampling from a fixed initialization distribution also. 

Therefore, in practice, choosing good initializations and architectures often reduces to experimentally testing several possibilities. While the need for experimentation can not be eliminated wholly, we propose a mechanism to test several weight initailizations and architectures, in lesser computational power. In this paper, we propose an early success indicator which predicts, at an early stage of training, the extent to which a DAI (dataset-architecture-initialization combination) that is trained using stochastic gradient descent would be successful. Such a score would be crucial to test several initializations and architectures with a reduced cost. By discarding DAIs that do not have the potential of learning effectively, at an earlier stage, we save computing power and run time. 

 We make the following contributions in this paper:
\begin{enumerate}
    \item Introduce early success indicators to predict the success of an initialization for a particular architecture and dataset. We also extend our framework to predict the success of different initializations and architectures  for a fixed dataset (Appendix \ref{a:2}).
    \item Analyze the performance of the proposed early success indicators through extensive empirical study on different datasets, architectures, and initializations. 
\end{enumerate}

The paper is organized as follows: Section \ref{s2} outlines the background work and formalisms required to build the scoring mechanisms. Section \ref{s3} introduces the early success indicators. Section \ref{s5} explains the experimental setting and section \ref{s6} states the evaluation technique for early success indicators. Section \ref{s7} presents the results, and section \ref{s8} establishes the conclusion and future work. 

\section{Background Work}\label{s2}
Trained neural networks are shown to exhibit compressibility property [\cite{Arora2018StrongerApproach}], which is exploited to design early success indicators.  
\subsection{Compressibility in trained neural networks}
The early success indicator introduced in this paper is motivated from the noise compression properties as outlined in [\cite{Arora2018StrongerApproach}]. 
In their paper, Arora et al. propose an empirical noise-stability framework for DNNs that computes each layer's stability to the noise injected at lower layers. For a well trained network, each layer’s output to an injected Gaussian noise at lower layers is stable. The added noise in a trained neural network (capable of generalization) gets attenuated as it propagates to higher layers.  This noise stability allows individual layers to be compressed.

  To test the noise stability of a network, a Gaussian noise in injected at one of the hidden layers and its propagation throughout the network is studied. It has been seen empirically that as noise propagates to deeper layers, the effect of noise diminishes. Noise sensitivity $\Psi_{N}$ of a mapping $M$ from a real-valued vector $x$ to a real-valued vector with respect to some noise distribution $N$ is defined as:
\begin{equation}
  \Psi_{N}(M,x) = \mathbf{E}_{\eta \in N}\left[\frac{||M(x+\eta||x||)-M(x)||^2}{||M(x)||^2}\right]  
\end{equation}

Noise stability from a linear mapping with respect to Gaussian noise has been derived mathematically in [\cite{Arora2018StrongerApproach}]. Their results prove that the noise sensitivity of a matrix $M$ at any vector $x \neq 0$ with respect to Gaussian distribution $N(0,I)$ is $\frac{||M||^2_F||x||^2}{||Mx||^2}$. This suggests that if a vector $x$ is aligned with matrix $M$, the noise sensitivity would be low, implying that the matrix $M$ would be less sensitive when noise is added at $x$. Low sensitivity implies that the matrix $M$ has some large singular values. This gives rise to some preferential directions along which signal $x$ is carried, while the noise, which is uniform across all directions, gets attenuated. Intuitively, this translates to a non-uniform distribution of singular values from the output of the transformation which generalizes well. On training further, as the noise stability of the neural network strengthens, the noise gets suppressed further, extent of compression increases, and the preferential direction of propagation gets aligned with the signal (higher singular values of $M$). 

It has been shown through experimentation that this compression property is observed for non-linear deep neural networks as well. We show that the singular values of the output matrix of hidden layers in a well trained DNN is non-uniform. We use this non-uniform distribution of singular values and its evolution as training continues to build the early success indicators and empirically show its utility in predicting the success of a DAI for feed-forward networks.

\subsection{Related Work} 
% NAS, hyperparameter search, stocasticity, optimizers
While there is active research being conducted on efficient hyperparameter optimization [\cite{Jasper2012}, \cite{Maclaurin2015}], neural architecture search [\cite{Liu2018DARTS:Search}, \cite{Zoph2016NeuralLearning}], learning to learn, and meta-learning [\cite{Andrychowicz2016}, \cite{Eggensperger2018}], the focus of these approaches is  on identifying good configurations, often requiring high computational power. This work, however, focuses on establishing a simple framework for making an early prediction on the success of a particular configuration of a DAI by facilitating early give-up for configurations which might not train well on further training.

\section{Early Success Indicator}\label{s3}

We utilize the preferential direction in which the signal is carried in well trained networks to predict the success of a DAI. We introduce two metrics to capture the singular value distribution of the output from hidden layers of DNNs at a training step and its evolution as training progress. Both of these metrics are combined to predict the success of a DAI early, enabling us to stop the training of networks that are likely not to benefit from training further.

\begin{figure*}[h]
\subfloat[ Training and Validation accuracy (DAI-I) \label{fig:1ia}]{\includegraphics[width=0.32\textwidth]{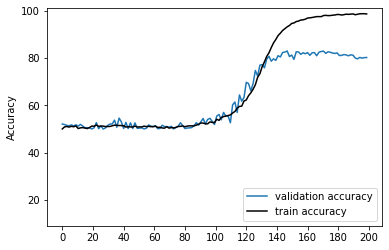}}\hfill
\subfloat[ Normalized singular values (sorted) - layer 1 of size 512 (DAI-I) \label{fig:1ib}]{\includegraphics[width=0.32\textwidth]{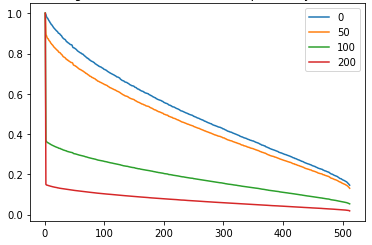}}\hfill
\subfloat[ Normalized singular values (sorted) - layer 2 of size 256 (DAI-II) \label{fig:1ic}]{\includegraphics[width=0.32\textwidth]{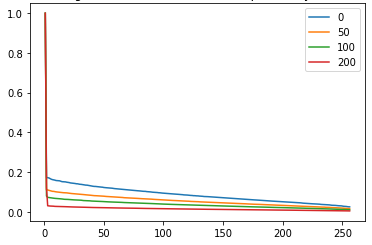}}\hfill
\subfloat[ Training and Validation accuracy (DAI-II) \label{fig:1id}]{\includegraphics[width=0.32\textwidth]{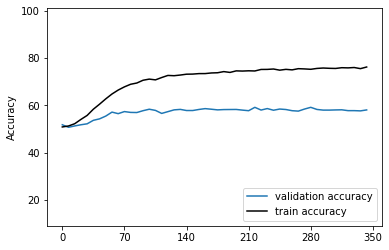}}\hfill
\subfloat[ Normalized singular values (sorted) - layer 1 of size 512 (DAI-II) \label{fig:1ie}]{\includegraphics[width=0.32\textwidth]{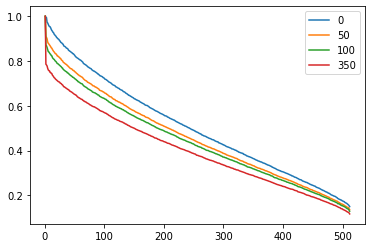}}\hfill
\subfloat[  Normalized singular values (sorted) - layer 2 of size 256 (DAI-II) \label{fig:1if}]{\includegraphics[width=0.32\textwidth]{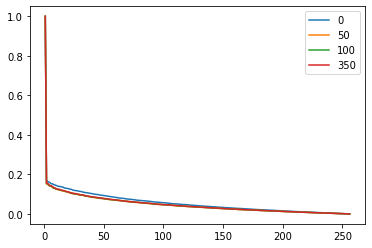}}\hfill
\caption{Evolution of singular values for \texttt{Shell} dataset, \texttt{4-layer MLP} architecture, initialization from \texttt{Normal Xavier} scheme in (\ref{fig:1ia})-(\ref{fig:1ic})  DAI-I trained till 200 epochs and (\ref{fig:1id})-(\ref{fig:1if}) DAI-II trained till 350 epochs.}
\label{fig:1}
\end{figure*}

We consider two Dataset-Architecture (DA) settings, and analyze the performance of the final learned network with different random initializations. For consistency of comparison, we have used the same training procedure (as detailed in Section \ref{s7} and trained on cross-entropy loss) that transforms the initialization parameters to the early success score parameters. This transformation is not entirely deterministic due to mini-batch effects, but we find these to be negligible for the synthetic datasets, and conjecture the same for larger datasets too. Figure \ref{fig:1} shows two initializations from one of the Dataset-Architecture combinations of \texttt{Shell} dataset (synthetic dataset consisting of two classes; detailed in Section \ref{s5}), 4-layer MLP architecture (layers of dimensions 512, 256, 256, 128 respectively). These initializations are trained using RMSProp optimizer with a learning rate of $10^{-3}$, and min-batch of size 32. The first initialization trains to a validation accuracy of 85.74\% at the end of 200 epochs. Figure \ref{fig:1ib}-\ref{fig:1ic} show normalized and sorted singular values obtained from the matrix of the outputs from the first two hidden layers of dimension 512 and 256 respectively (the largest singular value is normalized to 1). It can be seen that with each epoch, the distribution of singular value compresses. This can be contrasted with the performance of the second initialization. Even with training till 350 epochs, the model obtains a validation accuracy of only 55.42\%. The distribution of SVD values from the hidden layers (Figure \ref{fig:1ie}-\ref{fig:1if}) does not show significant compression as training progresses.

% Consider two different dataset, architecture, initialization (DAI) combination trained on the  same  dataset,  architecture,  training procedure, and initialization sampled from the same distribution: DAI-I and DAI-II. We train the two DAIs on , initialized from \texttt{Normal Xavier} scheme using RMSProp optimizer with a learning rate of $10^{-3}$ and trained on cross-entropy loss. Figure \ref{fig:1} shows that these networks with the same initialization distribution, architecture, and learning scheme could have very different generalization gaps and validation accuracy. DAI-I obtains a validation accuracy of 85.74\% at the end of 200 epochs. Figure \ref{fig:1ib}-\ref{fig:1ic} show normalized and sorted singular values obtained from the matrix of the outputs from the first two hidden layers of dimension 512 and 256 respectively (the largest singular value is normalized to 1). It can be seen that with each epoch, the distribution of singular value compresses. This can be contrasted with the performance of DAI-II. Even with training till 350 epochs, the model obtains a validation accuracy of only 55.42\%. The distribution of SVD values from the hidden layers (Figure \ref{fig:1ie}-\ref{fig:1if}) does not show significant compression as training progresses. 

Through extensive experimentation in the following sections, we claim that the insight obtained from this shifting distribution of singular values gives a prescience of the DAI's performance and thereby enables us to only train models with higher chances of generalizing well. We propose that deep neural networks capable of learning on further training are characterized by a steep decay of singular values as training progresses and quantify this characteristic to evaluate the utility of our indicators. 

{\textbf{Notation: }} We use the following notation for the task of multi-class classification with $k$ classes: Consider a dataset $\{(x_i,y_i)\}_{i=1}^n$, where $x_i \in \mathbb{R}^d$, label $y_i$ is an integer between 1 to $k$. A multi-class classifier $f$ transforms an input $x$ in $\mathbb{R}^d$ to $\mathbb{R}^k$. Let $f$ be a feed-forward neural network of depth $L$ parameterized by weight matrices $W_L,...,W_1$, where dimension of each hidden layer $i$ be $d_i$, such that $W_k \in \mathbb{R}^{d_k \times d_{k-1}} $. $(X,Y)$ be the test dataset with $m$ samples $\{(x_{ti},y_{ti})\}_{i=1}^m$. $A_i$ be the matrix obtained on passing the test dataset through the $i^{th}$ hidden layer of the trained network. $\{\sigma_{tij}\}_{j = 1}^{d_i}$ is the normalized singular values obtained in descending order from the matrix $A_i$ at epoch $t$. For a hidden layer $l$ at epoch $t$, we define two metrics $s_{olt}$ and $s_{slt}$. These matrices are computed over a window of the last $t_0$ epochs. 

\subsection{Capturing steep decay of the singular values - $s_{ot}$}
Metric $s_{ot}$ captures the non-uniform distribution of singular values at epoch $t$ and the preferential directions for signals which can compress noise in other directions. 
\begin{equation}
s_{olt} = \beta\times \frac{d_l \times t_0}{\sum_{k=t-t_0}^{t}\sum_{i=1}^{d_l}\sigma_{kli}}
\end{equation}

$s_{olt}$ is the score for output of layer $l$ at epoch $t$. An average score can be calculated across all layers as 

\begin{equation}
    s_{ot} = \frac{1}{L}\sum_{i=1}^{L}s_{olt}
\end{equation}
        
        $\beta$ is a hyper-parameter that is chosen empirically (set to $1$ for all experiments to follow).  The higher the value of $s_{ot}$, the higher the decay in the distribution of singular values. For a steep decay of singular values, $\sigma_{kli}$ for any $i > 1$  would result in small values with increasing $i$ (as the highest singular value is scaled to 1). This would result in a higher value of $s_{olt}$ for that layer. Therefore, $s_{ot}$ can capture the non-uniform distribution of singular values. A higher $s_{ot}$ suggests that the distribution of singular values is skewed. Models which generalize well show a high value of $s_{ot}$ during training.
    
\subsection{Capturing shift of SVD decay - $s_{st}$}
$s_{st}$ quantifies this automatic dimensionality reduction phenomenon across epochs. It captures the shift in the non-uniform distribution of singular values as training progresses. 
\begin{equation}
    s_{slt} = \frac{1}{(t_0-1)\times d_l}\times \sum_{k=t-t_0+1}^{t}\sum_{j=1}^{d_l}\alpha_{k}(\sigma_{(k-1)lj}-\sigma_{klj})
\end{equation}

 $s_{slt}$ is the score for output of layer $l$. An average score is calculated similarly as 
 \begin{equation}
    s_{st} = \frac{1}{L}\sum_{i=1}^{L}s_{slt} 
 \end{equation}

 $\alpha_k$ is a hyper-parameter that is chosen empirically. It is an increasing parameter with $k$. It varies linearly, giving highest weightage to the latest epoch and least weightage to the initial epoch of the window. $\alpha_k$ lies in the range $[0,1]$.
 The higher the value of $s_{st}$, the more the shift towards decay. As the validation loss of a model reduces, its transformation matrix gets aligned with the input data leading to further skewness and positive value in $s_{st}$.

The scorings $s_{st}$ and $s_{ot}$ are normalized across different feed-forward architectures by normalizing the score for each hidden layer by the number of units in the layer. Parameter $\beta$ is used to scale $s_{ot}$ in the range of $s_{st}$, which is crucial when combining the two scores.  These two metrics are combined, and a scoring is proposed, independent of the labels of the validation dataset, which can be used as an early indicator of predicting the success of a DAI trained using gradient descent.

For a particular DAI combination, an early success indicator calculates the score by combining the two metrics $s_{ot}$ and $s_{st}$ at every epoch $t$. This combined score can be used to make a binary decision on whether to continue training on that DAI or discard this particular choice. If the score crosses a certain threshold $t_1$, the model is trained further. However, if the score is below $t_1$, the model is discarded before completing the training procedure. This can allow several initializations to be tested for different datasets and tasks with lesser time and computational power. We calculate the final score by combining the two metrics:
\begin{equation}
    s_t = |log (s_{ot})| + \eta \times s_{st}
\end{equation}

$\eta$ is a hyper-parameter used to take a weighted average of the two terms in the combined score (chosen to be $3.5 \times 10^3$). It is worth noting that compressed singular values does not necessarily imply a subspace in which the signal propagates effectively. There can be configurations of neural networks with extremely compressed singular values, that do not optimize or generalize further on training. Such configurations are captured through $s_{st}$. To discard such models which do not train further, we introduce two more thresholds $t_2$ and $\delta$. These thresholds allow us to remove models which have compressed SVD values and still show very small changes in $s_{st}$. This phenomenon is usually accompanied by non-decreasing validation loss.  Training these models further does not change the distribution of singular values and therefore the training could be stopped earlier. Considering these factors, the final combined score $s_F$ is summarized as follows: 

\begin{equation}
  s_F =
    \begin{cases}
     0  & \text{if } \text{$s_{ot}$} \geq t_2 \text{ and $s_{st}$} \leq \delta   \\
       s_t& \text{otherwise}
    \end{cases}       
\end{equation}

The score $s_F$ is calculated on validation data points but does not require the labels of the data points. However, the scores can be improved with the availability of labels for the validation dataset. We incorporate the validation accuracy (normalized in the range [0, 1]) at a particular epoch $t$ as $v_t$ into the scoring mechanism and propose a new score $s_{Fv}$ at that epoch as follows:

\begin{equation}
  s_{Fv} =
    \begin{cases}
     v_t & \text{if } \text{$s_{ot}$} \geq t_2 \text{ and $s_{st}$} \leq \delta   \\
       v_t \times (1+ min(\frac{1-v_t}{v_t}, \gamma s_t ))& \text{otherwise}
    \end{cases}       
\end{equation}

The $min$ function ensures that the score $s_{Fv}$ lies in the interval $[0,1]$. It is an indicative measure of the final validation accuracy of a DAI combination based on the current validation accuracy $v_t$. $\gamma$ is a hyper-parameter to scale the score $s_t$, and is constant for all DAIs ($\gamma$ is taken as $2 \times 10^{-4} \times v_t \times t$, where $t$ is the current epoch at which the scores are calculated).

\section{Experimental Setting}\label{s5}

The goal of early success indicators $s_{F}$ and $s_{Fv}$ is to predict which DAIs would be successful. In the following sections, we confine the analysis of early success indicators to testing various initializations for a fixed task, dataset, architecture, and training mechanism. We show an extension of the set-up to testing combinations of different architectures and initializations on a fixed task, dataset and training mechanism in Appendix \ref{a:2}. We define a "good" initialization as the weight matrices which yield a good validation accuracy upon training on a fixed dataset and architecture using a fixed training procedure. Obtaining a good neural network model is heavily affected by the choice of initial parameters. Initial point has an influence on whether or not the learning process converges. In cases where it converges, initialization can affect how fast or slow the convergence happens, the error, and the generalization gap. There is no universal initialization technique as modern initialization strategies are heuristic. The current understanding of how initial point affects optimization and generalization is still incipient, and therefore, in practice, choosing good initialization is hugely experimental. To determine the success of initialization on a particular architecture and dataset, the model needs to be trained for several epochs (till either the loss reaches a certain threshold or overfitting begins). This process consumes significant compute power and time and can limit the number of initializations to be tried out. Therefore, this framework can, to a reasonable extent, determine the success of a DAI during the early training process and provide insights on the benefits of further training. These scores can aid in making an informed decision on the benefits of further training the model, thereby saving computing power and time by stopping the training of DAIs that do not have the potential of learning effectively. 

\begin{figure*}[h]
    \centering
    \subfloat[ DA-I \label{fig:1c}]{\includegraphics[width=0.42\textwidth]{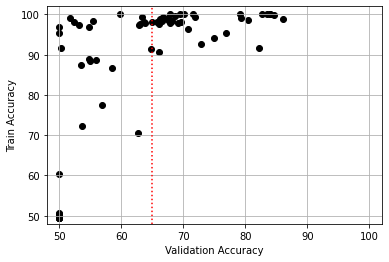}}
    \subfloat[ DA-II \label{fig:1c}]{\includegraphics[width=0.42\textwidth]{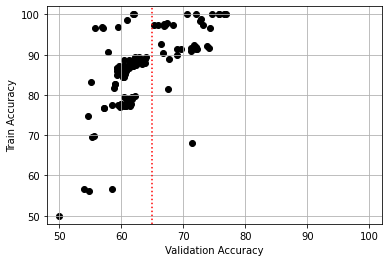}}
    \caption{Distribution of the final training and validation accuracy for (a) DA-I and (b) DA-II at epoch 100.}
\label{fig:2a}
\end{figure*}

 The extent of success of a DNN model, in our set-up, is taken as the final validation accuracy (accuracy of validation data in the last epoch $e_L$). The early success indicators, which are calculated at an earlier epoch $e_C $ $(e_C < e_L)$ (called as \textit{checkpoint epoch}), give a score which is indicative of the success of the DAI. This score enables us to identify and discard models that would not train sufficiently at the end of training and continue training the remaining. A higher positive value of score indicates higher chances of success of the DAI. We evaluate the performance of early success indicators on classification task.

\subsection{Dataset} 
Experiments have been conducted on synthetic and real-world datasets:  \texttt{Shell}, \texttt{CIFAR-10} (\ref{a:1}) for classification.

\begin{figure}[h]
    \centering
    \includegraphics[width=0.25\textwidth]{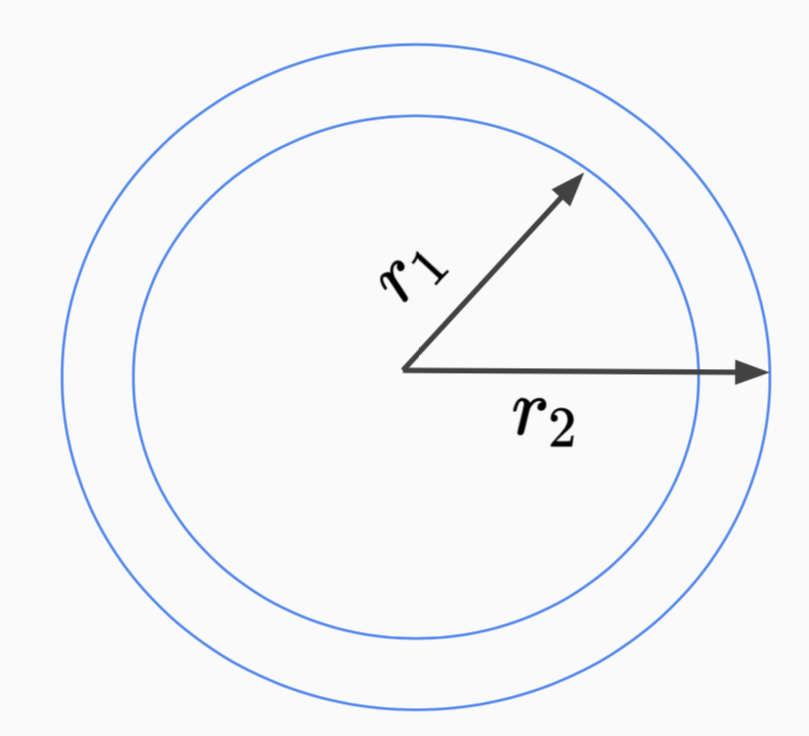}
    \caption{Visualization of the distribution of \texttt{Shell} dataset in ${R}^2$. The two classes lie on shells of different radius $r_1$ and $r_2$ respectively.}
    \label{fig:3_ex_1}
\end{figure}

\texttt{Shell} dataset is a synthetic dataset comprising of 2 classes. Each class lies in a shell in $ \mathbb{R}^d$. $d$ is taken to be 1024 by default. In the experiments, the two classes form a concentric shell of radius 1 and 1.1 respectively. 20,000 training samples and 4,000 validations samples are used throughout the experiments. In 2-dimensional space ($d = 2$), \texttt{Shell} data can be visualized on the 2 curves as shown in Figure \ref{fig:3_ex_1}. 

\subsection{Architecture} The experiments in this section have been done on the fully connected layers of multi-layer perceptron (MLP). Preliminary experiments done on the fully connected layers in convolutional neural network architectures (CNNs) have also shown success (refer Appendix \ref{a:1}).

\subsection{Initialization } 
Different initializations are generated by randomly varying the standard deviation of the weights in Normal Xavier initialization within a factor of 10. Please note that, throughout the paper, an initialization scheme refers to initializing the weights of neural network layers by sampling from a specific distribution (for example \texttt{Normal Xavier} initialization scheme). An initialization, on the other hand, refers to the specific weight matrices sampled from an initialization scheme.

 In all the experiments, rectified linear activation function (ReLU) activation is used. Optimizer used is one of RMSProp, Adam. Loss function is cross-entropy for classification. We further demonstrate the robustness of early success indicators by training on shuffled labels, thereby generating models with varied generalization gaps. The network is trained on a dataset, $y\%$ of whose labels have been randomly shuffled (where $y \in (0, 100)$) (Refer Appendix \ref{a:3}).

\section{Evaluation}\label{s6}

Checkpoints are defined as epochs at which the scores are calculated and a decision regarding the success of the initialization is made. The decision for a DAI can be one among: discarding the models that would not train or stopping further training for models that are trained (but training further won't yield further benefits) and continuing training models which show potential for better results. This decision is taken only based on the values from $s_F$ (and $s_{Fv}$ when validation labels are available) at the checkpoint epoch. To evaluate the performance of these scoring, all initializations are further trained further till a pre-decided final epoch and the prediction at the checkpoint epoch is compared against the validation accuracy at the final epoch. The early success indicator score for an initialization is computed by training a neural network initialized from that initialization using stochastic gradient descent algorithm, fixing the optimizer and its corresponding optimization parameters. At each checkpoint epoch, we pass the test data through the hidden layers of the neural network and obtain its singular values using SVD. These singular values are sorted and normalized across each hidden layer to calculate the scores. While we fix the algorithm that translates the initialization weights to the final score, this algorithm could also be a source of randomness due to mini-batch gradient. However, we observe in the \texttt{Shell} dataset that success is primarily decided by the initialization, i.e. a given initialization is either a success or a failure once we fix the algorithm regardless of the mini-batch randomness.
\begin{figure*}[h]
\subfloat[ Validation accuracy at each epoch for the five initializations \label{fig:1c}]{\includegraphics[width=0.3\textwidth]{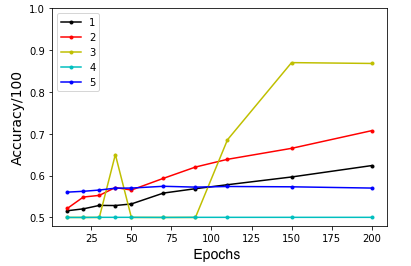}}\hfill
\subfloat[ $s_F$  at each epoch for the five initializations \label{fig:1c}]{\includegraphics[width=0.3\textwidth]{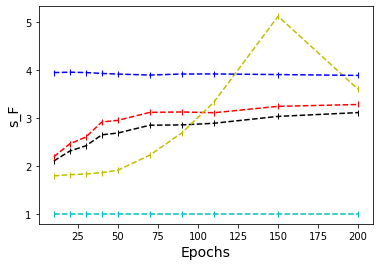}}\hfill
 \subfloat[ $s_{Fv}$ at each epoch for the five initializations \label{fig:1c}]{\includegraphics[width=0.3\textwidth]{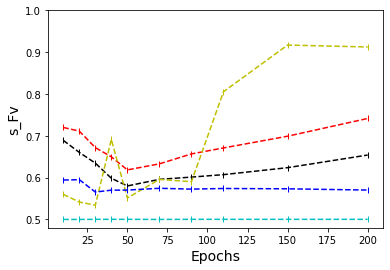}}\hfill
\caption{Variation of scores $s_F$ and $s_{Fv}$ with epochs and their corresponding validation accuracy for five random initializations of DA-1. } \label{fig:3a}
\end{figure*}

Evaluation is done using two metrics: Spearman's rank correlation coefficient and the number of correct predictions for the success of a DAI. We calculate the Spearman's correlation between the score obtained from early predictor at checkpoint epoch and validation accuracy at the final epoch to evaluate how accurately the early success predictor can predict the performance of the model.

 \textbf{Spearman's Rank Correlation Coefficient:} It evaluates the monotonic relationship between two continuous or ordinal variables. In a monotonic relationship, the variables tend to change together, but not necessarily at a constant rate. The Spearman's correlation coefficient is based on the ranked values for each variable rather than the raw data. It is defined as the Pearson Correlation Coefficient between the rank variables. Mathematically, for two variables $X$ and $Y$, it is given as:
\begin{equation}
r_s = \rho_{rg_{X}, rg_{Y}} =\frac{cov(rg_{X},rg_{Y})}{\sigma_{rg_{X}}\sigma_{rg_{Y}}}  
\end{equation}
The notations are as follows:
\begin{itemize}
    \item $r_s$ is the Spearman's Rank Correlation coefficient
    \item $rg_{X}$ and $rg_{Y}$ are ranks of $X$ and $Y$
    \item $cov$ is the covariance
    \item $\sigma_{rg_{X}}$ is the variance of the rank variable of $X$
\end{itemize}

Correlation coefficients vary between -1 and +1, with 0 implying no correlation. Correlations of -1 or +1 imply a strong positive relationship. Positive correlations imply that as $x$ increases, so does $y$. Negative correlations imply that as $x$ increases, $y$ decreases. We compute the Spearman's rank correlation between the scores calculated at the checkpoint epoch and the final validation accuracy to show that the two share a positive relation. A high positive correlation indicates that the scoring is a good predictor for the final validation accuracy. We compute Spearman's correlation between $s_{Fv}$ at a checkpoint epoch and the final validation accuracy. In order to test the efficacy of this score, we compare its performance with the correlation of validation accuracy at checkpoint epoch with the final validation accuracy as the baseline.

Scores $s_F$ and $s_{Fv}$ can also be evaluated by formulating the prediction of success of gradient descent for a particular DAI as a two-class classification problem. For a fixed dataset, architecture, and learning scheme, we test several initializations. An arbitrary threshold is set for all DAs to mark successful initializations. Initializations that achieve final validation accuracy above a desired threshold belong in class $y_1$ of successfully trained models. Initializations that do not attain this threshold upon full training belong to class $y_0$ of not successfully trained models. A prediction is made at the checkpoint epoch to predict the class in which the model trained of that initialization would belong in the final epoch.

\section{Results}\label{s7}

We consider the following DAs for evaluation:
\begin{itemize}
    \item \textbf{DA-I}: 120 initializations trained on \texttt{Shell} dataset (1024 dimensional input vectors), MLP architecture with four hidden layers of dimensions 512, 256, 256, 128 respectively. Optimizer used is Adam with learning rate $10^{-4}$, and mini-batch size as 32. All initializations are trained to a total of 100 epochs to evaluate the performance of the metrics. Initializations that achieve validation accuracy above 65\% at the final epoch are considered to be in class $y_1$ and all the other initializations in class $y_0$. On final training, we get 32 out of the 120 models in class $y_1$ and 88 in class $y_0$. 
    
    \item \textbf{DA-II}: 137 initializations trained of \texttt{Shell} dataset (256 dimensional input vectors), MLP architecture with two hidden layers of dimensions 256, 128 respectively. Optimizer used is RMSprop with learning rate $10^{-4}$, and mini-batch size as 32. All initializations are trained to a total of 100 epochs to evaluate the performance of the metrics. Initializations that achieve validation accuracy above 65\% at the final epoch are considered to be in class $y_1$ and all the other initializations in class $y_0$. On final training, we get 34 out of the 137 models in class $y_1$ and 103 in class $y_0$.
\end{itemize}

Figure \ref{fig:2a} shows the distribution of final training accuracies and validation accuracies for the 120 initialization of DA-I and the 137 initialization of DA-II at epoch 100. The points right to the red line indicates models which qualify to be in class $y_1$. 
 \begin{figure*}[h]
 \begin{tabular}{l|l|c|c|c}
\multicolumn{1}{c}{}&\multicolumn{1}{c}{}&\multicolumn{2}{c}{Prediction}&\\
\cline{3-4}
\multicolumn{1}{c}{}&\multicolumn{1}{c|}{}&$y_1$&$y_0$&\multicolumn{1}{c}{Total}\\
\cline{2-4}
\parbox[t]{2mm}{\multirow{2}{*}{\rotatebox[origin=c]{90}{True}}}& $y_1$ & $8$ & $24$ & $32$\\
\cline{2-4}
& $y_0$ & $7$ & $81$ & $88$\\
\cline{2-4}
\multicolumn{1}{c}{} & \multicolumn{1}{c}{Total} & \multicolumn{1}{c}{$15$} & \multicolumn{1}{c}{$105$} & \multicolumn{1}{c}{$120$}\\
\multicolumn{5}{c}{(a) Prediction using $s_{F}$} \\
\end{tabular}
\begin{tabular}{l|l|c|c|c}
\multicolumn{1}{c}{}&\multicolumn{1}{c}{}&\multicolumn{2}{c}{Prediction}&\\
\cline{3-4}
\multicolumn{1}{c}{}&\multicolumn{1}{c|}{}&$y_1$&$y_0$&\multicolumn{1}{c}{Total}\\
\cline{2-4}
\parbox[t]{2mm}{\multirow{2}{*}{\rotatebox[origin=c]{90}{True}}}& $y_1$ & $20$ & $12$ & $32$\\
\cline{2-4}
& $y_0$ & $10$ & $78$ & $88$\\
\cline{2-4}
\multicolumn{1}{c}{} & \multicolumn{1}{c}{Total} & \multicolumn{1}{c}{$30$} & \multicolumn{    1}{c}{$90$} & \multicolumn{1}{c}{$120$}\\
\multicolumn{5}{c}{(b) Prediction using $s_{Fv}$} \\
\end{tabular}
\begin{tabular}{l|l|c|c|c}
\multicolumn{2}{c}{}&\multicolumn{2}{c}{Prediction}&\\
\cline{3-4}
\multicolumn{2}{c|}{}&$y_1$&$y_0$&\multicolumn{1}{c}{Total}\\
\cline{2-4}
\parbox[t]{2mm}{\multirow{2}{*}{\rotatebox[origin=c]{90}{True}}}& $y_1$ & $18$ & $14$ & $32$\\
\cline{2-4}
& $y_0$ & $10$ & $78$ & $88$\\
\cline{2-4}
\multicolumn{1}{c}{} & \multicolumn{1}{c}{Total} & \multicolumn{1}{c}{$28$} & \multicolumn{1}{c}{$92$} & \multicolumn{1}{c}{$120$}\\
\multicolumn{5}{c}{(c) Prediction using validation accuracy}  \\
\end{tabular}
\caption{Breakdown of the prediction of eventual success of all initializations in DA-I using $s_F$, $s_{Fv}$ and validation accuracy at checkpoint epoch 10.} \label{fig:5.8}
\end{figure*}

\begin{figure*}
\begin{tabular}{l|l|c|c|c}
\multicolumn{2}{c}{}&\multicolumn{2}{c}{Prediction}&\\
\cline{3-4}
\multicolumn{2}{c|}{}&$y_1$&$y_0$&\multicolumn{1}{c}{Total}\\
\cline{2-4}
\parbox[t]{2mm}{\multirow{2}{*}{\rotatebox[origin=c]{90}{True}}}& $y_1$ & $19$ & $15$ & $34$\\
\cline{2-4}
& $y_0$ & $7$ & $96$ & $103$\\
\cline{2-4}
\multicolumn{1}{c}{} & \multicolumn{1}{c}{Total} & \multicolumn{1}{c}{$26$} & \multicolumn{1}{c}{$111$} & \multicolumn{1}{c}{$137$}\\
\multicolumn{5}{c}{(a) Prediction using $s_F$}  \\
\end{tabular}
\begin{tabular}{l|l|c|c|c}
\multicolumn{1}{c}{}&\multicolumn{1}{c}{}&\multicolumn{2}{c}{Prediction}&\\
\cline{3-4}
\multicolumn{1}{c}{}&\multicolumn{1}{c|}{}&$y_1$&$y_0$&\multicolumn{1}{c}{Total}\\
\cline{2-4}
\parbox[t]{2mm}{\multirow{2}{*}{\rotatebox[origin=c]{90}{True}}}& $y_1$ & $25$ & $9$ & $34$\\
\cline{2-4}
& $y_0$ & $0$ & $103$ & $103$\\
\cline{2-4}
\multicolumn{1}{c}{} & \multicolumn{1}{c}{Total} & \multicolumn{1}{c}{$25$} & \multicolumn{    1}{c}{$112$} & \multicolumn{1}{c}{$137$}\\
\multicolumn{5}{c}{(b) Prediction using $s_{Fv}$} \\
\end{tabular}
\begin{tabular}{l|l|c|c|c}
\multicolumn{2}{c}{}&\multicolumn{2}{c}{Prediction}&\\
\cline{3-4}
\multicolumn{2}{c|}{}&$y_1$&$y_0$&\multicolumn{1}{c}{Total}\\
\cline{2-4}
\parbox[t]{2mm}{\multirow{2}{*}{\rotatebox[origin=c]{90}{True}}}& $y_1$ & $18$ & $16$ & $34$\\
\cline{2-4}
& $y_0$ & $0$ & $103$ & $103$\\
\cline{2-4}
\multicolumn{1}{c}{} & \multicolumn{1}{c}{Total} & \multicolumn{1}{c}{$18$} & \multicolumn{1}{c}{$119$} & \multicolumn{1}{c}{$137$}\\
\multicolumn{5}{c}{(c) Prediction using validation accuracy}  \\
\end{tabular}
\caption{Breakdown of the prediction of eventual success of all initializations in DA-II using $s_F$, $s_{Fv}$ and validation accuracy at checkpoint epoch 10.} \label{fig:5n.11}
\end{figure*}

 Figure \ref{fig:3a} shows the value of validation accuracy, $s_F$, and $s_{Fv}$ at every epoch for five randomly selected initializations from DA-I. Score $s_F$ and $s_{Fv}$ give an estimate at each epoch on the success of further training the model. The models with a horizontal line in the second curve (initialization 4 and 5) represent models whose score does not change much. While $s_F$ can determine that their accuracies might not change with further training, it can not determine the actual accuracy. This limitation is mitigated when using $s_{Fv}$. When using the validation labels, $s_F$ can determine that the training has saturated and $s_{Fv}$ can predict the final validation accuracy to be close to the current validation accuracy $v_t$ itself (might even reduce with over-fitting).

At checkpoint epoch, the points with $s_F$ greater than $t_1 = 0.25$ are predicted to train successfully for all DAs. We evaluate the performance of proposed scores on both the dataset-architecture combinations.

\subsection{Results on DA-I}

Table \ref{tab:5.8} compares the performance of $s_F$, $s_{Fv}$, and validation accuracy at three different checkpoint epochs. Correlation of $s_{Fv}$ at checkpoint epoch 10 and 20 and final validation accuracy is higher than that of the validation accuracy at checkpoint epochs and the final validation accuracy. $s_{Fv}$ and validation accuracy at checkpoint epoch 30 have high correlation values.
% The correlation plots of the four different measure - train accuracy, $s_F$, validation accuracy, $s_{Fv}$ at checkpoint epoch 30 and final validation accuracy at epoch 100 is shown in Figure \ref{fig:5.09a}.

\begin{table}[h]
\centering
 \begin{tabular}{||c c c c||} 
 \hline
 Checkpoint &$s_F$& $s_{Fv}$& Validation accuracy   \\ 
  epoch &at checkpoint& at checkpoint  &  at checkpoint \\ [0.5ex] 
 \hline\hline
 10 & 0.404 &\textbf{0.707} & 0.685 \\ 
 \hline
 20 &  0.773 &\textbf{0.829} & 0.820 \\ 
 \hline
 30  &0.831 & 0.872 &\textbf{0.878}\\ \hline
 \end{tabular}
 \caption{Correlation of $s_F$, $s_{Fv}$,  validation accuracy at different checkpoint epochs with the final validation accuracy for DA-I.}\label{tab:5.8}
\end{table}

Using the validation accuracy at the checkpoint epoch to predict the validation accuracy at the final epoch can lead to several mis-classifications. A low validation accuracy can either stay at low validation accuracy or get well-trained as well. A high validation accuracy can also either yield a lower final validation accuracy (over-fitting) or get better trained. Score $s_{Fv}$ supplements this with additional information, boosting the validation accuracy score that has the potential to train further and a low $s_{st}$ to indicate stopping the training process. Figure \ref{fig:5.8} summarizes the comparison of $s_F$, $s_{Fv}$, and validation accuracy at checkpoint epoch 10 on predicting the class ($y_0$ or $y_1$) of the initializations. It can be seen that score $s_{Fv}$ is better than the baseline of validation accuracy in the classification problem.

\subsection{Results on DA-II}

Similar to DA-I, a prediction is made whether  the  model,  after  getting trained till 100 epochs, would belong in class $y_1$ or $y_0$ using these scores at earlier epochs. Figure \ref{fig:5n.11} compares the performance of $s_F$, $s_{Fv}$, and validation accuracy at the checkpoint epoch 10 to make a binary prediction of the success of initializations. It can be seen that $s_{Fv}$ is better at predicting the success of the initializations, compared to simply using the validation accuracy for this DA.

 Table \ref{tab:5n.51} summarizes the Spearman's correlation values obtained for all the initializations at different checkpoints. It can be concluded that score $s_{Fv}$ is better than the baseline of validation accuracy at the checkpoint at all checkpoints in this case as well.
\begin{center}
\begin{table}[h]
\centering
 \begin{tabular}{|| c c c c||} 
 \hline
 Checkpoint  & $s_F$   & $s_{Fv}$& Validation accuracy   \\ 
 epoch  &   at checkpoint & at checkpoint &   at checkpoint \\ [0.5ex]
 \hline\hline
 10  &0.395 & \textbf{0.766} & 0.716 \\ 
 \hline
 20 &0.324& \textbf{0.767} & 0.762 \\ 
 \hline
 30  &0.414& \textbf{0.818} & 0.786 \\ [1ex]
 \hline \end{tabular}
 \caption{Correlation $s_F$, $s_{Fv}$,  validation accuracy at different checkpoint epochs with the final validation accuracy for all initializations in DA-II.}\label{tab:5n.51}
\end{table}
\end{center}

It must be noted that Spearman's correlation is not an ideal metric to evaluate the performance of score $s_F$. Score $s_F$ can be low for models which have already saturated and would not benefit any more from further training, irrespective of the validation accuracy they saturate at.

\section{Conclusion}\label{s8}

In this work, a mechanism has been realized to predict the success of gradient descent for a particular dataset-architecture-initialization combination. Two early success indicators are proposed, analyzed, and tested on different combinations of datasets, architectures, and initializations to facilitate early give-up, i.e., stopping the training of models early, which are predicted to not generalize well upon further training, thereby saving computational time and power. These scores can be used to supplement the available validation accuracy at an early epoch to make predictions about the future success of the model.

The utility of success indicators can be further extended by exploiting the labels of the validation data points. The calculation of $s_{st}$ and $s_{ot}$ can be computed differently for individual classes. Intuitively, it is expected that the amplifying power of the matrix obtained from the data points of one class would be in different directions than the amplifying power of matrix obtained from data points of a different class. If it were in the exact same direction, distinction between the two classes could not be made out. Another natural extension of the work would be test the effectiveness of the proposed metrics on larger models, exploring different tasks, architectures, and datasets. 
\newpage

% \begin{figure*}[h]
% \subfloat[ $s_F$ and final validation accuracy \\
% \label{fig:1c}]{\includegraphics[width=0.49\textwidth]{Plots/Correlation_Sf.png}}\hfill
%  \subfloat[ Train accuracy at epoch 30 and final validation accuracy\\  \label{fig:109c}]{\includegraphics[width=0.45\textwidth]{Plots/Correlation_train.png}}\hfill
% \subfloat[ $s_{Fv}$ and validation accuracy \\
% \label{fig:1c}]{\includegraphics[width=0.49\textwidth]{Plots/Correlation_Sfv.png}}\hfill
%  \subfloat[ Validation accuracy at epoch 30 and 200 \\  \label{fig:1c}]{\includegraphics[width=0.45\textwidth]{Plots/Correlation_val.png}}\hfill
% \caption{$s_F$, training accuracy, $s_{Fv}$,  validation accuracy at checkpoint epoch 30 plotted against the final validation for all initializations in DA-I.} \label{fig:5.09a}
% \end{figure*}

% \section*{Acknowledgment}

% The authors would like to thank...

	\bibliographystyle{apalike}
\bibliography{bare_conf}
% \bibitem{IEEEhowto:kopka}
% H.~Kopka and P.~W. Daly, \emph{A Guide to \LaTeX}, 3rd~ed.\hskip 1em plus
%   0.5em minus 0.4em\relax Harlow, England: Addison-Wesley, 1999.

\newpage
\clearpage

\section*{Appendix}

\subsection{Early success indicators applied to Convolutional Neural Networks}%
\makeatletter\def\@currentlabel{A.1}\makeatother
\label{a:1}

\begin{figure}[h]
\subfloat[ Validation accuracy = 76.3\% \\at epoch 100\\
\label{fig:1c}]{\includegraphics[width=0.23\textwidth]{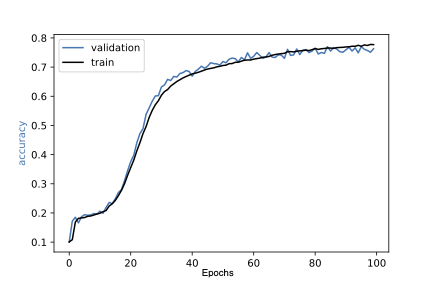}}
 \subfloat[Normalized  singular  values  (sorted)  -  Dense layer of size 256\\  \label{fig:1c}]{\includegraphics[width=0.23\textwidth]{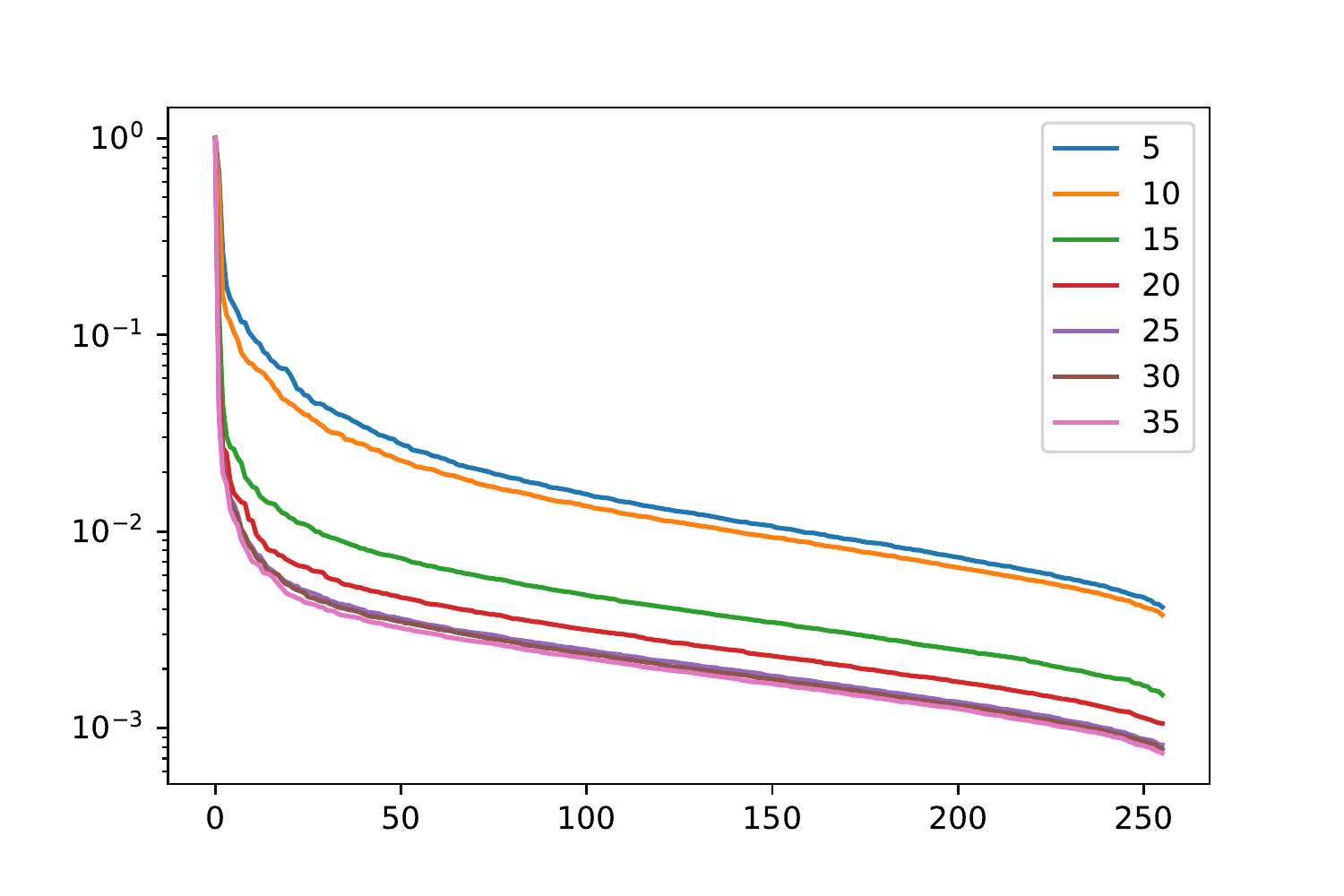}}\hfill
 \subfloat[ Validation accuracy = 10.5\%\\ at epoch 100\\ 
\label{fig:1c}]{\includegraphics[width=0.23\textwidth]{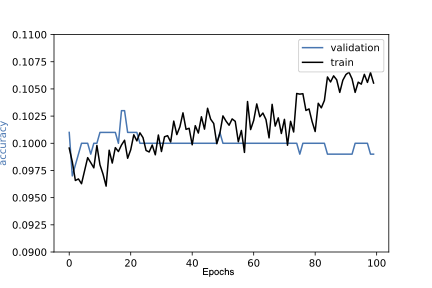}}\
\subfloat[Normalized  singular  values  (sorted)  -  Dense layer of size 256\\
\label{fig:1c}]{\includegraphics[width=0.23\textwidth]{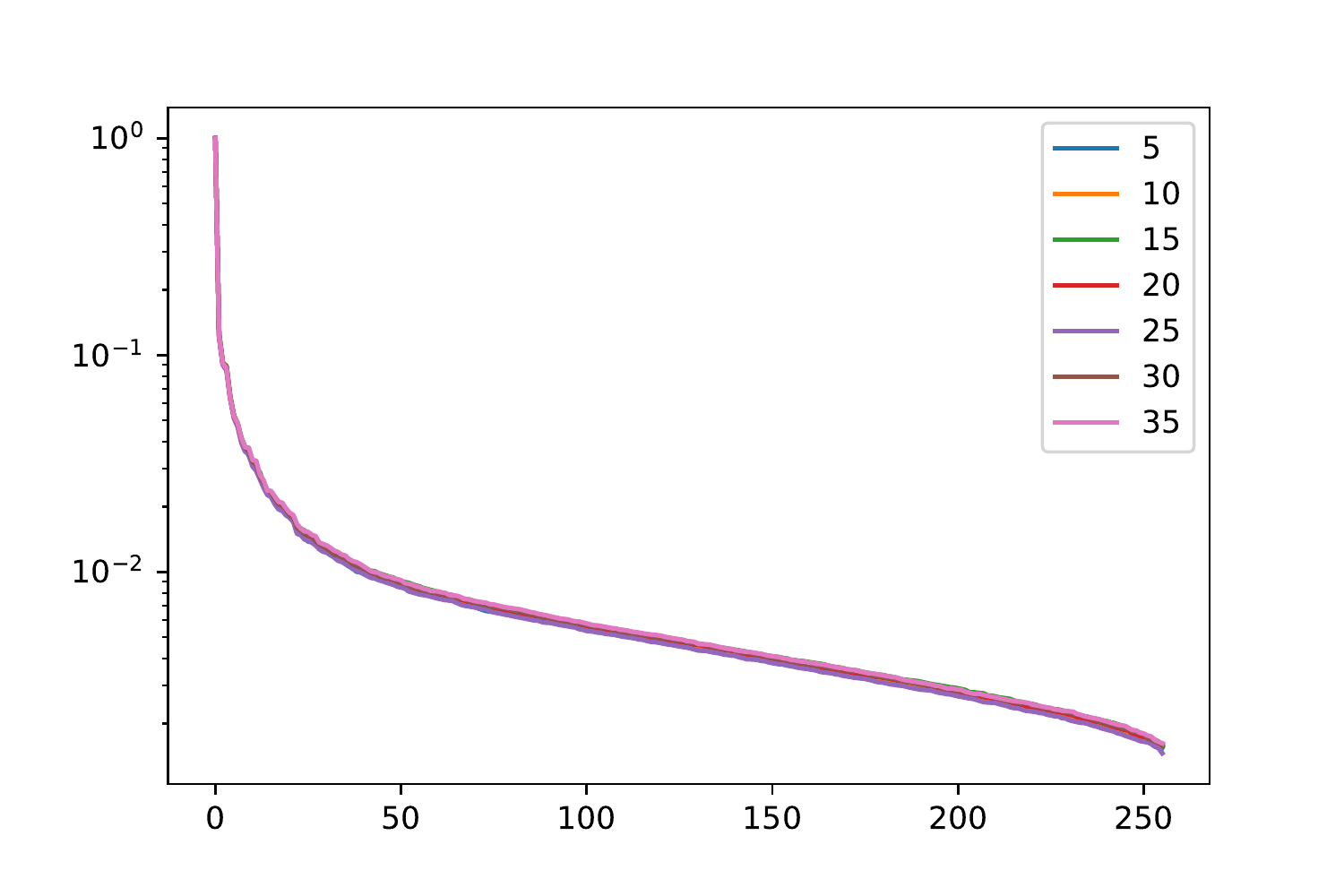}}\
 \hfill
\caption{Dense layer in well trained CNN show a decay in the singular values upon training on $CIFAR-10$.} \label{fig:3m}
\end{figure}

Obtaining singular values of the convolutional layers in CNN requires concatenating the vector obtained from passing the data through all the feature maps in the layer. This method can require large memory and computation. Although this can be computed approximately (using pooling, sampling, etc), we show that the dense layers in CNNs shows a compression in SVD values (as observed in MLP) upon training. Consider two initializations trained on \texttt{CIFAR-10}, \texttt{10-layer CNN}  ($C16 - A - C16 - A - M - C32 - A - D - C32 - A - M - C64 - A - C64 - A - C128 - A - M -  F256 - A - F128 - A - F10 -softmax$, where $C$ is Conv2D layer, $A$ is activation (ReLU), $M$ is MaxPooling, $F$ is Fully connected Layer, $D$ is dropout) in Figure \ref{fig:3m}. It can be seen that the first model trains till a validation accuracy of 76.3\% on training till 100 epochs. This is accompanied by the compression of singular values in the fully connected layer of dimension 256 as training continues. The second initialization does not show this compression and remains untrained till epoch 100. 

However, it is observed that the range of $s_{st}$ and $s_{ot}$ for CNNs is different from that in MLP. Hence, a better normalization technique is required for the early success indicators to analyze the results obtained from the SVD of fully connected layers in CNNs. 

\subsection{Performance of early success indicators across architecture}%
\begin{figure}[h]
    \centering
    \includegraphics[width = 0.4\linewidth]{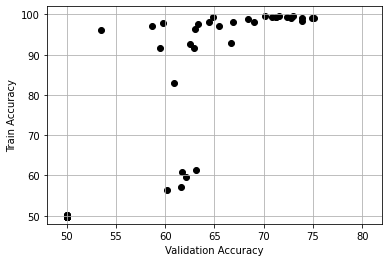}
    \caption{ Final training and validation performance of 50 different architectures on \texttt{Shell-1024} dataset.}
    \label{fig:sd}
\end{figure}

\makeatletter\def\@currentlabel{A.2}\makeatother
\label{a:2}
The analysis so far has been limited to the same dataset and architecture. However, finding an optimal DNN architecture is also largely experimental. This section expands the utility of early success indicators to test the prediction of both an initialization and architecture for a dataset. For the experiment, \texttt{Shell-1024} dataset is considered. The various MLP architecture chosen vary from 1 hidden layer to 11 hidden layers, and the dimension of each hidden layer is chosen arbitrarily in the range of 32 to 512 (in factors of 2). All the models are trained on Adam optimizer with a learning rate of $10^{-4}$. A total of 50 such architectures and initializations are trained on this dataset, and their $s_F$ and $s_{Fv}$ values are recorded at the checkpoint epoch. All the models are further trained for 100 epochs to find the final validation accuracy. The distribution of final training and validation accuracy for all the models can be observed in Figure \ref{fig:sd}.

The correlation values at checkpoint 10, 20, 30 are summarized in Table \ref{tab:5n.59}. $s_{Fv}$ beats the validation accuracy at checkpoint for all the checkpoints. 
\begin{center}
\begin{table}[h]
\centering
 \begin{tabular}{||c c c c||} 
 \hline
Checkpoint &$s_F$ &$s_{Fv}$& Validation accuracy   \\ 
 epoch  &  at checkpoint & at checkpoint &  at checkpoint \\ [0.5ex] 
 \hline\hline
 10  & 0.529 &\textbf{0.719} & 0.671 \\ 
 \hline
 20  & 0.632 & \textbf{0.789} & 0.752 \\ 
 \hline
 30  & 0.657 &\textbf{0.843} & 0.814 \\ [1ex]
 \hline \end{tabular}
 \caption{Correlation of $s_F$, $s_F$, $s_{Fv}$,  validation accuracy at different checkpoint epochs with the final validation accuracy for all 50 architectures trained on \texttt{Shell} dataset.}\label{tab:5n.59}
\end{table}
\end{center}

\subsection{Performance of early success indicators on shuffled labels}%
\begin{figure*}[h]
\centering
\subfloat[ Training and Validation accuracy on correct labels \label{fig:1a}]{\includegraphics[width=0.32\textwidth]{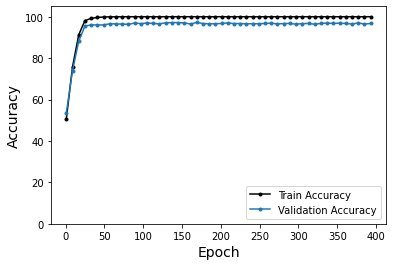}}\hfill
\subfloat[ Normalized  singular  values  (sorted)  -  Layer 1 of size 128 (Correct labels) \label{fig:1a}]{\includegraphics[width=0.32\textwidth]{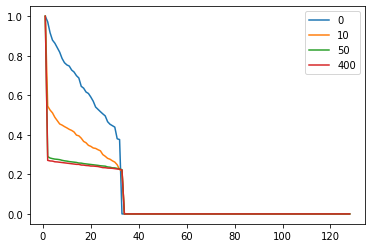}}\hfill
\subfloat[ Normalized  singular  values  (sorted)  -  Layer 1 of size 128 (Correct labels) \label{fig:1a}]{\includegraphics[width=0.32\textwidth]{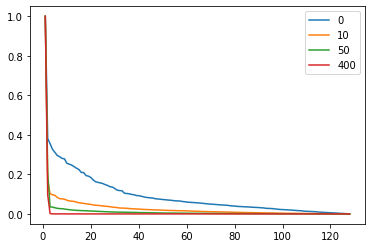}}\hfill
% \subfloat[  20\% Shuffled \label{fig:1c}]{\includegraphics[width=0.32\textwidth]{Images/SVD/prelim/228.png}}\hfill
\subfloat[  Training and Validation accuracy on 40 \% Shuffled \label{fig:1c}]{\includegraphics[width=0.32\textwidth]{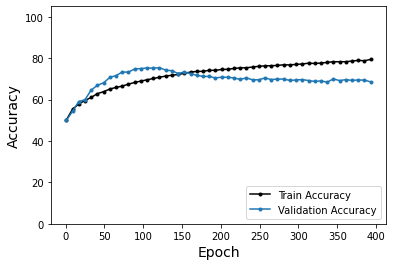}}
\subfloat[Normalized  singular  values  (sorted)  -  Layer 1 of size 128  (40 \% Shuffled) \label{fig:1c}]{\includegraphics[width=0.32\textwidth]{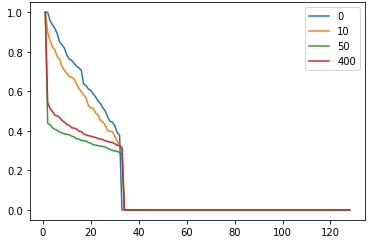}}\hfill
\subfloat[ Normalized  singular  values  (sorted)  -  Layer 1 of size 128  ( 40 \% Shuffled) \label{fig:1c}]{\includegraphics[width=0.32\textwidth]{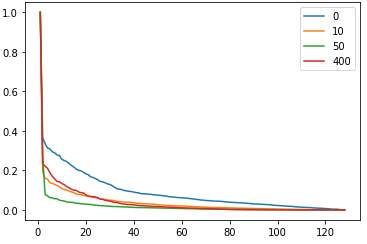}}\hfill
\caption{SVD output for (Shuffled) \texttt{Shell} dataset (32 dimensional).} \label{fig:4.8}
\end{figure*}

\makeatletter\def\@currentlabel{A.3}\makeatother
\label{a:3}
\cite{Zhang2016UnderstandingGeneralization}  experimentally showed that the standard architectures using SGD and regularization could reach low training error even when trained on randomly labeled examples (which clearly won’t generalize). In this experiment, 40\% of labels are shuffled for \texttt{Shell} dataset (32 dimensional). Architecture consists of two hidden layers of dimension $\{128,128\}$ and initialization from \texttt{Normal Xavier} scheme. Optimizer used is SGD with a learning rate of $10^{-3}$. The results are presented in Figure \ref{fig:4.8}. 

In this experiment, when the labels are shuffled
the validation accuracy is higher than the training accuracy for few epochs but eventually goes below training accuracy with further training. Interestingly, the singular value plot for each layer also displays such variation. The values were initially decaying but with further training it expands again. The singular value decomposition starts reverting back as the validation accuracy dips further, resulting in a negative $s_{st}$. This phenomenon can be crucial in detecting overfitting in models in the absence of validation labels. This demonstrates the robustness of SVD based scores on networks that reach almost zero training error but have a huge generalization gap.

% that's all folks
\end{document}